\documentclass{ecai}  

\usepackage{graphicx}
\usepackage{latexsym}
\usepackage{caption}
\usepackage{subcaption}
\usepackage{booktabs}

\usepackage{bbding}
\usepackage{pifont}
\usepackage{threeparttable}
\usepackage{amssymb}
\usepackage{amsmath}
\usepackage{multirow}
\usepackage{booktabs}
\usepackage{array}
\usepackage{hyperref}
\usepackage{float} 
\usepackage{stfloats}

\begin{document}

\begin{frontmatter}

\title{Dynamic Graph Convolutional Network with 
Attention Fusion for Traffic Flow Prediction}
\author[A]{\fnms{Xunlian}~\snm{Luo}}
\author[B]{\fnms{Chunjiang}~\snm{Zhu}}
\author[A]{\fnms{Detian}~\snm{Zhang}\thanks{Corresponding Author. Email: detian@suda.edu.cn.}}
\author[C]{\fnms{Qing}~\snm{Li}}

\address[A]{Institute of Artificial Intelligence, School of Computer Science and Technology, Soochow University, China.}
\address[B]{Department of Computer Science, University of North Carolina at Greensboro, USA.}
\address[C]{Department of Computing, The Hong Kong Polytechnic University, Hong Kong, China.}
\address{20215227114@stu.suda.edu.cn, chunjiang.zhu@uncg.edu, detian@suda.edu.cn, qing-prof.li@polyu.edu.hk}

\begin{abstract}
Accurate and real-time traffic state prediction is of great practical importance for urban traffic control and web mapping services. With the support of massive data, deep learning methods have shown their powerful capability in capturing the complex spatial-temporal patterns of traffic networks. However, existing approaches use pre-defined graphs and a simple set of spatial-temporal components, making it difficult to model multi-scale spatial-temporal dependencies. In this paper, we propose a novel dynamic graph convolution network with attention fusion to tackle this gap. The method first enhances the interaction of temporal feature dimensions, and then it combines a dynamic graph learner with GRU to jointly model synchronous spatial-temporal correlations. We also incorporate spatial-temporal attention modules to effectively capture long-range, multifaceted domain spatial-temporal patterns. We conduct extensive experiments in four real-world traffic datasets to demonstrate that our method surpasses state-of-the-art performance compared to 18 baseline methods.
\end{abstract}

\end{frontmatter}

\section{Introduction}
In urban computing, popular web mappings services such as Google Maps, and Bing Maps heavily rely on accurate traffic flow prediction as a backend for their frontend applications such as personalized route planning and closest restaurant/gas station recommendation. Similarly, departments of transportation in next-generation smart cities often need to allocate traffic resources and optimize traffic control plans, {e.g.}, for an exposition, by blocking a minimum number of streets while avoiding serious traffic congestion. The plan needs to be generated in advance and thus requires precise traffic flow estimation. Recently, we have witnessed the bloom of online ride-hailing/sharing services, {e.g.}, Uber, Lyft, and Didi. They often provide functions to request a ride on the Web directly with no need to download apps. These Web-based services, again, need a powerful traffic prediction engine. The traffic flow prediction model analyzes large amounts of historical traffic data to estimate future traffic conditions based on statistical or data-driven methods \cite{tedjopurnomo2020survey}, providing a basis for decision-making for these applications. However, traffic flow is affected by complex spatial and temporal dependencies, which make accurate real-time traffic forecasting extremely challenging.

\begin{figure}[!t]
\centering
\includegraphics[width=84mm]{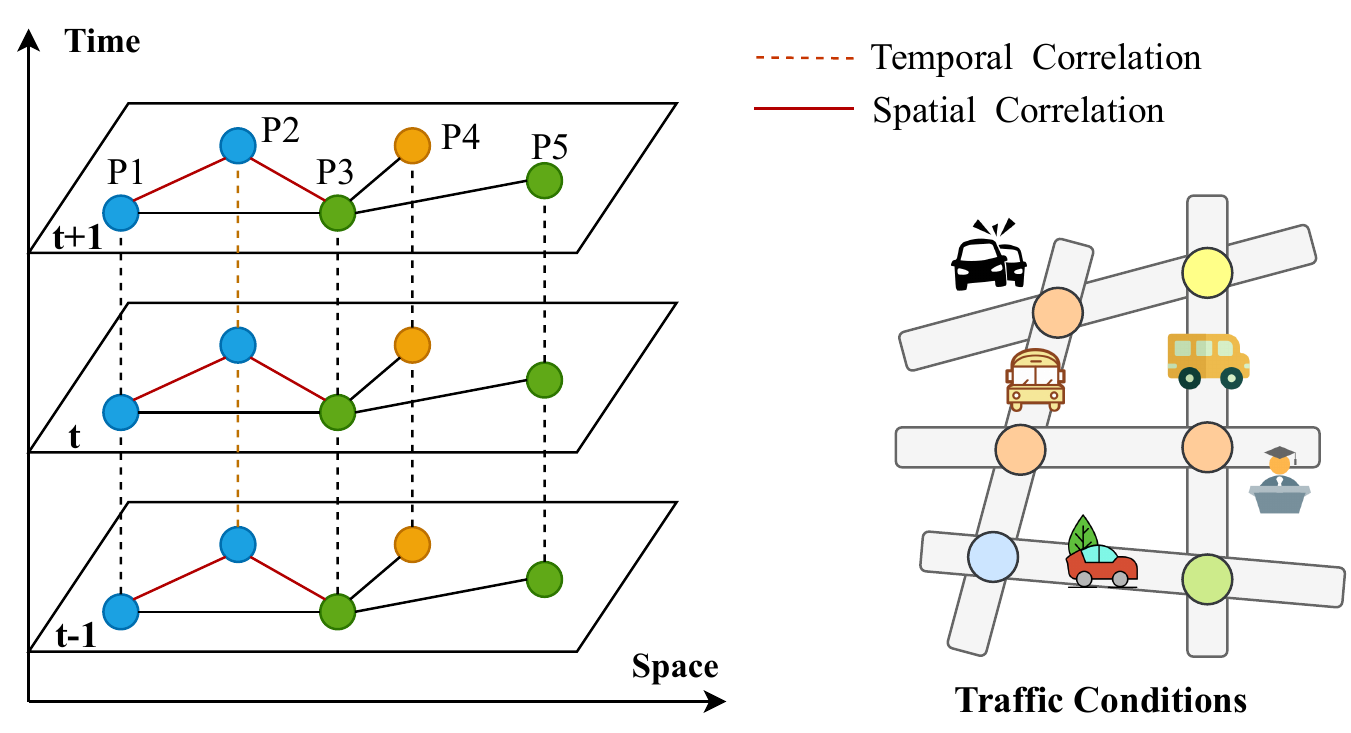}
\caption{Complex spatial-temporal correlations.}
\label{fig:label1}
\vspace{-2mm}
\end{figure}

As shown in Fig. \ref{fig:label1}, traffic prediction differs from traditional time series in that it has significant spatial and temporal correlation and is also susceptible to other external factors \cite{DBLP:journals/eswa/JiangL22}. The variation of traffic in time has proximity and trend. It is also influenced by the travel routine of urban residents, which shows a certain periodicity over a long period of time. Being connected by a non-Euclidean spatial road network, the traffic volume at the observation point is subject to the propagation of vehicles from neighboring roads upstream and downstream, exhibiting mobility and transience. Besides, the intervention of some external factors (weather, graduation ceremony, traffic accidents, etc.) also can cause drastic fluctuations in traffic flow. Additionally, the spatial and temporal correlations are typically multi-scale, which can be roughly divided into global and local in space and long-term and short-term in time.

Extensive research has been carried out to address these challenges. Statistical models ({e.g.}, VAR \cite{lu2016integrating}, ARIMA \cite{kumar2015short}) and machine learning methods including SVR \cite{DBLP:journals/tits/WuHL04} and ANN \cite{van2012short}, treat traffic sequences as independent data streams, considering only the temporal variability of the data and ignoring spatial correlation. The rapid development of deep learning techniques brings traffic prediction research to a new stage. In particular, CNN \cite{DBLP:journals/corr/abs-1803-01271}, RNNs (its variants LSTM, GRU) \cite{DBLP:journals/corr/ChungGCB14}, and Transformer with self-attention \cite{DBLP:conf/nips/VaswaniSPUJGKP17} have been successful in modeling sequence tasks. A novel attempt is to use convolution neural networks to model the implicit spatial relationships of urban regions as they do for images or videos \cite{DBLP:conf/aaai/ZhangZQ17}. Later studies identify the advantages of graph neural networks over CNNs in extracting structure information of irregular topologies \cite{DBLP:journals/tnn/WuPCLZY21}. Inspired by the spatial-temporal characteristics of traffic data, an intuition naturally develops to combine graph convolution networks with sequence models to jointly model spatial-temporal dependencies.

Spatial-temporal graph neural networks provide a feasible solution for traffic prediction. Some models such as DCRNN \cite{DBLP:conf/iclr/LiYS018} combines diffusion convolution with GRU-based encoder-decoder architecture to capture the directionality of traffic propagation. STSGCN \cite{DBLP:conf/aaai/SongLGW20} proposes a local spatial-temporal graph to represent the transfer of information in the temporal and spatial directions, and adopts residual-connected GCN modules to model spatial-temporal heterogeneity. However, they ignore the fact that pre-defined graphs cannot fully reflect the reality of traffic connections. Although GWNET \cite{DBLP:conf/ijcai/WuPLJZ19} and MTGNN \cite{DBLP:conf/kdd/WuPL0CZ20} improve their practices with adaptive graph learning, the separated spatial-temporal components are unable to capture synchronized correlations. GMAN \cite{DBLP:conf/aaai/ZhengFW020} based on spatial-temporal attention fusion shows promising results for long-term forecasting but performs poorly in the short term. STG-NCDE \cite{choi2022graph} and DSTAGNN \cite{lan2022dstagnn} achieve good prediction performance, but bring large computational consumption and system overhead. These works have contributed to a tremendous improvement in performance from multiple perspectives. However, we observe that they are not combined in a cohesive way to address comprehensive features and their correlations, and rarely balance method performance and complexity. 

To address the shortcomings of existing methods, in this paper, we propose a Spatial-Temporal \textbf{A}ttention \textbf{F}usion \textbf{D}ynamic \textbf{G}raph \textbf{C}onvolution \textbf{N}etwork (\textbf{AFDGCN}) for traffic flow prediction. Specifically, We employ feature augmentation, dynamic graph convolution recurrent networks, and multiple attention fusion to capture complex traffic patterns from multi-scale spatial-temporal aspects. In summary, the main contributions of this paper are as follow: 

\begin{itemize}
\item A novel and efficient spatial-temporal framework is proposed for traffic prediction. The traffic features are first augmented to improve the information interaction of the data on the feature channel and the temporal axis, rather than performing a basic channel ascending dimension.

\item We propose a dynamic graph learner for capturing hidden spatial dependencies and combine adaptive graph convolution with gated recurrent units to build a dynamic graph convolution recurrent network for capturing local spatial-temporal correlations.

\item We consolidate temporal attention and spatial attention to improve the model's ability that identifies changes in the spatial field of view and long-range time dependence. This indeed reduces the error of overall prediction in complex scenes and enhances the robustness of the model.

\item We conduct extensive experiments in four benchmark datasets, and evaluate the performance compared with 18 baseline methods. The results show that our proposed method outperforms all other methods in three standard evaluation metrics with low computation costs. 
\end{itemize}

\section{Related Work}
\subsection{Graph Convolution Networks}
Graph convolution networks process data as a set of graph structures consisting of nodes and edges \cite{DBLP:journals/corr/HenaffBL15}, and have a wide range of applications in social network analysis, recommender systems and molecular biology. Its core idea is to aggregate the features of a node with those of its neighboring nodes to generate a new feature representation of the node \cite{DBLP:journals/tnn/WuPCLZY21}. Generally, there are two mainstream graph convolution networks, namely, the spectral-based methods and the spatial-based methods. Chebyshev graph convolution network (ChebNet) is a representative spectral method that transforms the graph signal into the frequency domain for filtering to achieve noise reduction and feature extraction \cite{DBLP:conf/nips/DefferrardBV16}. Specifically, by computing the Chebyshev polynomial $T_k(x)$ to approximate the filter, a Laplacian matrix $\widetilde{L}$ can be obtained as follows:
\begin{equation}
\widetilde{L} = \frac{2}{\lambda_{max}}L-I_{N} = \sum_{k=0}^{K-1}\theta_kT_k(\widetilde{L})
\end{equation}
 where $L = I_N - D^{-\frac{1}{2}}AD^{-\frac{1}{2}}=U \Lambda U^{T}$ is the normalized graph Laplacian matrix, $U$ is the eigenvalue matrix of $L$. 
 
\noindent The ChebNet is expressed as follows:
\begin{equation}
\Theta \star_{\mathcal{G}} X=\Theta(L) X= \sigma\left(\sum_{k=0}^{K-1}\theta_kT_k(\widetilde{L})x_i\right)
\end{equation}
where $X \in \mathbb{R}^{N \times d}$ is the feature matrix of all nodes in the graph, $A \in \mathbb{R}^{N \times N}$ is the adjacency matrix of the graph and $k$ is the order of the ChebNet.

In traffic prediction, graph convolutional networks (GCNs) are generalized to high-dimensional GCNs by approximation of first-order Chebyshev polynomials \cite{DBLP:conf/iclr/KipfW17} as follows:
\begin{equation}
X^{(l)}=(I_N + D^{-\frac{1}{2}}AD^{-\frac{1}{2}})X^{(l-1)}W + b.
\end{equation}
where $D$ is the degree matrix of $A$, $W$ and $b$ are learnable weights and biases, and $X^{(l)}$ denotes the hidden representation of the $l$-layer.

\subsection{Deep Learning for Traffic Prediction}
A basic assumption behind spatial-temporal graphs for traffic forecasting is that the future state of a node depends on its historical knowledge and the information of its neighbors \cite{DBLP:conf/ijcai/WuPLJZ19}. It is natural to combine graph convolution networks with sequence models to construct spatial-temporal framework for modeling the spatial-temporal dependence of traffic data. The models can be classified as CNN-Based, RNN-Based and Attention-Based according to their temporal components \cite{DBLP:conf/cikm/JiangYWWDLCDSS21}. The CNN-based approaches use dilation convolution to expand the receptive field of the model with high computational efficiency, but the fixed implicit time step sacrifices some flexibility. RNNs provide a universal architecture of encoder-decoder. Many works \cite{DBLP:conf/iclr/LiYS018, DBLP:journals/corr/abs-2104-14917} replace the fully connected layer of GRU with GCN to achieve synchronous spatial-temporal modeling. Compared to the previous two, the attention-based methods \cite{DBLP:conf/aaai/ZhengFW020, DBLP:conf/aaai/GuoLFSW19} are relatively flexible and can learn to go further back in time patterns. The ongoing advancement of these efforts has made deep learning the primary method for spatial-temporal data mining. Meanwhile, the study of graph quality becomes a more general task. This has a direct impact on whether the model can effectively capture spatial dependencies. Adaptive graphs \cite{DBLP:conf/nips/0001YL0020} and discrete sampling graphs \cite{DBLP:conf/ijcai/YuLYLHWL22} in many works have been shown to model information propagation in space better than pre-defined graphs. Compared to analyzing temporal correlations, it is much more challenging to model spatial relations in traffic with insufficient prior knowledge.

\section{Problem Statement}
Traffic forecasting refers to the use of historically observed traffic data to predict the traffic state in a future period based on road network knowledge. The traffic sensor distribution is represented as a graph $\mathcal G = (V, E, A)$. $V$ is the set of $N=|V|$ nodes in the spatial topology and $E$ is the set of edges connected by node pairs. $A \in \mathbb{R}^{N \times N}$ is the adjacency matrix of the graph $\mathcal G$, indicating the proximity of connections or distances based on the Gaussian kernel function. Denote the traffic flow observed on $\mathcal G$ at time step $t$  as a graph signal matrix $X_t \in \mathbb{R}^{N \times C}$,  where $C$ is the feature channel ({e.g.} flow, speed, demand) of each node. 

Similarly, $X_{t-T:t-1} \in \mathbb{R}^{T \times N \times C}$ denotes the traffic statistics of the full road network at time intervals of $T$. The aim of traffic forecasting is to learn a function $f$ that is able to forecast $Q$ future signal tensor given $P$ historical signal tensor and the graph $\mathcal G$:
\begin{equation}
\label{problem}
[X_{t-P-1}, \cdots, X_{t}; \mathcal G] \underset{\Theta}{\stackrel{f}\longrightarrow} [\widehat{X}_{t+1}, \cdots ,\widehat{X}_{t+Q}]
\end{equation}
where $\Theta$ denotes the model parameters to be optimized.

\section{Proposed Model}
\begin{figure*}[t]
\centering
\includegraphics[width=\linewidth]{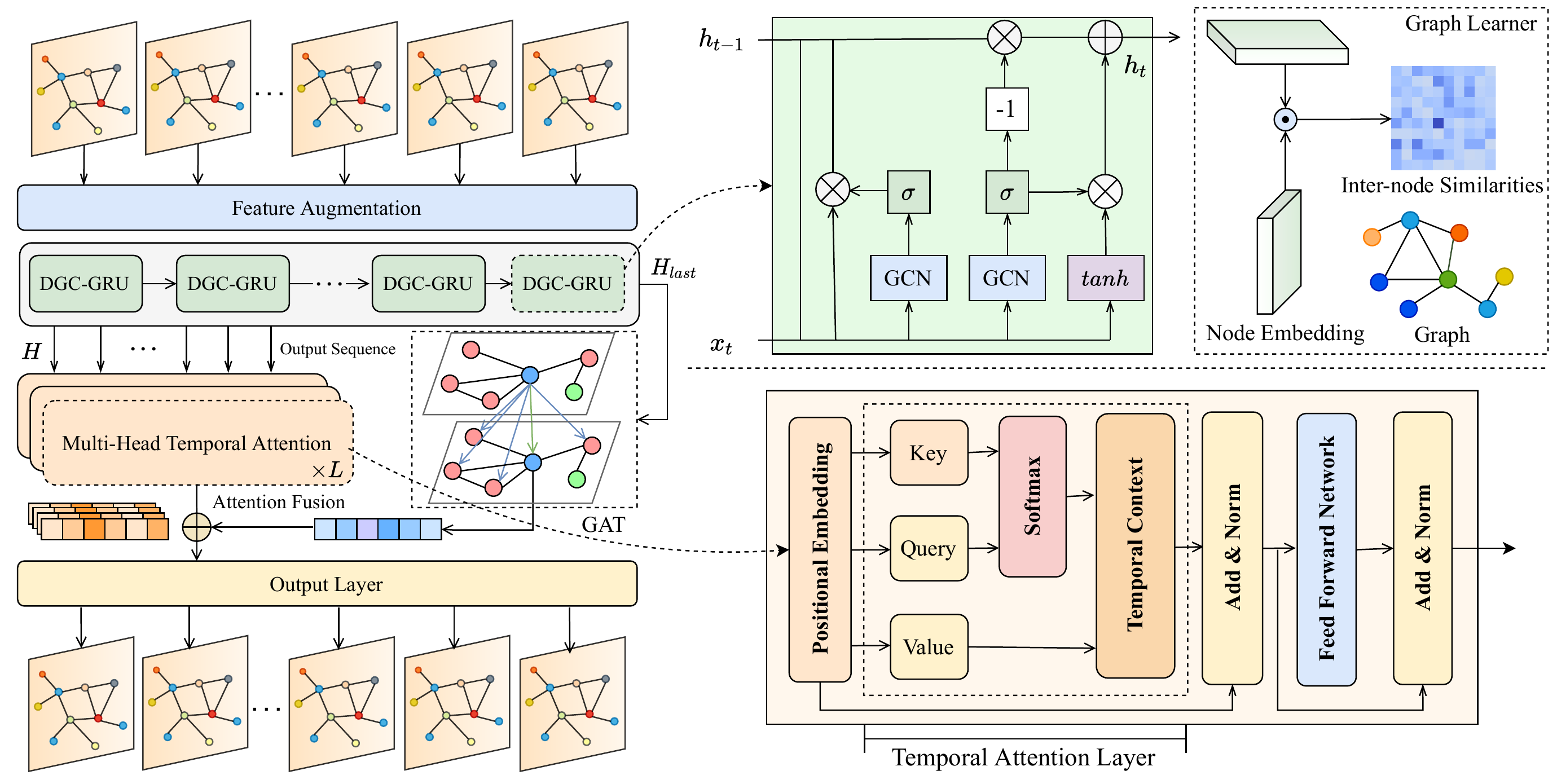}
\caption{The model overview of dynamic graph convolution network with spatial-temporal attention fusion.}
\label{fig:model}
\end{figure*}

We propose a novel spatial-temporal graph neural network called AFDGCN.  It consists of four main modules, namely feature augmentation layer, dynamic graph convolutional recurrent network, multi-head temporal attention module, and graph attention module. The structure of the model is illustrated in Fig. \ref{fig:model}.

\subsection{Feature Augmentation Layer}
To gain a comprehensive understanding of the intrinsic characteristics of time series data, we develop a feature augmentation layer that is inspired by channel-wise attention \cite{woo2018cbam}. The layer is based on a "squeeze-excitation" structure \cite{DBLP:abs-2112-05561} that enhances the information interaction between the channels and the temporal dimension. It improves the expressiveness of the feature map by automatically calibrating the importance of the feature space.

Specifically, the feature augmentation layer consists of two sub-structures in series, namely the channel calibration block and the temporal calibration block. The channel block is a fully connected network with two layers, which first performs a squeezing operation on the feature map $X \in \mathbb{R}^{T \times N \times D}$. This helps the model recalibrate the features with a low-dimensional distribution. It then analyzes the influence of each channel through a single activation process and a self-threshold mechanism. The channel-wise computation is formalized as follows:
\begin{equation}
s^{(c)}=\sigma(W_2\delta(W_1X^{(c)} + b_1) + b_2),
\end{equation}
\begin{equation}
X_1^{(c)}= X^{(c)} \circ s^{(c)}.
\end{equation}
where $W_1, W_2$ and $b_1, b_2$ are the weight parameters and bias of the fully connected layer. $\circ$ is the Hamada product. $\sigma$ and $\delta$ are the Sigmoid and ReLU activation functions, respectively.

Similarly, to focus on the structural information (significant time points in $P$) of the time series, the temporal calibration block utilizes a temporal convolution layer for feature extraction. After channel calculation and dimensional exchange, the feature map $X_1$ served as the input to the temporal-wise block. That is:
\begin{equation}
s^{(t)}=\sigma(\Theta_{1,k} \star \delta(\Theta_{1,k} \star X_1^{(t)}) ,
\end{equation}
\begin{equation}
X_2^{(t)}= X_1^{(t)} \circ s^{(t)}.
\end{equation}
where $\Theta_{1,k}$ is the $1 \times k$ convolution kernel. After preprocessing by feature augmentation layer, the reweighted feature maps are used as the output to feed in the training of the subsequent structures.

\subsection{Dynamic Graph Convolutional Recurrent Network}
The performance of traditional GCNs relies on a pre-defined graph structure. However, the connections built during the study based on distance proximity are often incomplete and biased \cite{DBLP:conf/ijcai/WuPLJZ19, DBLP:conf/kdd/WuPL0CZ20}. And the nodes' own factors ({e.g.}, POI, road structure, and road type) are difficult to represent \cite{DBLP:journals/tkde/GuoLWLC22}. These make general GCNs unable to capture spatial dependencies effectively. In view of these problems, we propose a dynamic graph learner for mining the potential spatial properties of traffic data. It generates the similarity matrix by initializing the node embeddings and the GCN takes into account the unique patterns of each node during feature aggregation. The parameters are dynamically updated during the training stage to adapt to the traffic data in an end-to-end manner.

All nodes are predefined with the embedding vector $E_{G}\in \mathbb{R}^{N\times d}$, where $d$ denotes the node embedding matrix, $d \ll N$. Then, we infer the implicit correlation between  node pairs by $E_G\cdot E^T_G\in \mathbb{R}^{N\times N}$. The normalized adjacency matrix is defined as:
\begin{equation}
\widetilde{A} = D^{-\frac{1}{2}}AD^{-\frac{1}{2}}=softmax(ReLU(E_G \cdot E^T_G)),
\end{equation}
where $E_G^{(i)} \in \mathbb{R}^d$ is the node embedding vector of $v_i$. The ReLU activation function is used for matrix sparsification and to eliminate the effect of negative values.

We observe that the weight and bias used by the traditional GCN for feature linear transformation are shared by all nodes. This ignores the differences in the traffic patterns of the nodes themselves. 
Inspired by the matrix decomposition,  we perform matrix multiplication of $W_{G} \in R^{d \times C \times F}$ ($b_{G} \in R^{d \times F}$) with the node embedding $E_G \in R^{N \times d}$ to generate new model parameters $W = E_GW_{G}$ ($b = E_Gb_{G}$, respectively). In this way, the dynamic generation of graph convolution $\Theta_x \star \mathcal G $ is expressed as:
\begin{equation}
Z=(I_N+softmax(ReLU(E_G \cdot E^T_G)))XE_GW_G+E_Gb_G.
\end{equation}

To capture synchronous temporal and spatial dependencies, we extend the fully connected operation in GRU with GCN to obtain the Dynamic Graph Convolutional Recurrent Network (DGCGRU). Due to the superior performance of GRU in handling sequential tasks, we embed graph convolution operations in the original GRU structure to capture sequential information while merging spatial relationships. Structurally, in addition to the present data input, each time step receives a hidden representation from the previous time step \cite{DBLP:conf/iclr/LiYS018}, which is used to control the memorization and forgetting of information. Here, the GRU operation is applied to each node in the graph and the parameters are shared with each other.

Specifically, given the previously hidden representation $H_{t-1}$ and the input data $X_{:,t}$ at time step $t$, we represent the computation of a single-step gated recurrent unit as: 
\begin{equation} 
\begin{aligned}
z_t &= \sigma\left(\Theta_z \star \mathcal G[X_{:,t},H_{t-1}]+b_z\right) \\
r_t &= \sigma\left(\Theta_r \star \mathcal G[X_{:,t},H_{t-1}]+b_r\right) \\
\hat{H}_t &= tanh\left(\Theta_h \star \mathcal G[X_{:,t}, (r \odot H_{t-1})]+b_{h}\right) \\
H_t &= z \odot H_{t-1} + (1-z) \odot \hat{H}_t,
\end{aligned} \label{equations}
\end{equation}
where $b_z,b_r$ and $b_h$ are learnable parameters of the recurrent neural network and $H_t$ is the hidden output at time $t$.

The combination of dynamic graph convolution and GRU enhances the model's ability to handle fine-grained spatial-temporal patterns. This dynamically generated graph can be freed from the constraints of pre-defined graphs because it adaptively builds connection relationships of node pairs using training data. However, GRU does not rigidly memorize all fixed-length sequences, but selectively stores information of historical time steps through hidden states due to the forgetting characteristic of gating. 
Although it can effectively capture local temporal information from traffic sequence data, it is insensitive to long-distance temporal relationships.

\subsection{Multi-head Temporal Attention Module}
A multi-head temporal attention module is used to extract the global context information of the sequence \cite{DBLP:conf/www/Wang0WJWTJY20, DBLP:journals/corr/abs-2001-02908}. This module consists of a positional embedding, a multi-head temporal attention layer, and a residual network. The temporal attention enables the model to observe longer-term temporal trends and focus on highly correlated information to compensate for the deficiencies of GRU in modeling long-range temporal dependence.

The hidden layer sequence $\left\{H_1[i, :], H_2[i, :], \cdots, H_T[i, :]\right\}$ from DGCGRU is fed to the temporal attention module. Given that the computation of the multi-head attention mechanism \cite{DBLP:conf/nips/VaswaniSPUJGKP17} ignores the relative positions of the sequences, we add a position encoding to the feature data at each time step. That is:
\begin{equation}
\hat{H_t}[i,:]=H_t[i,:] + e_t, \text{where} \ H_t \in \mathbb{R}^{N \times D}
\end{equation}
where the position token $e_t$ is defined as:
\begin{equation}
e_t=\begin{cases}
\sin (t/10000^{2i/d_{model}}) ,& \text { if } \ t=0, 2, 4, \ldots \\
\cos (t/10000^{2i/d_{model}}),& \text{otherwise}
\end{cases}
\end{equation}

The embedded features $\hat{H} \in \mathbb{R}^{T \times N \times D}$ are first projected into a high-dimensional latent subspace to generate the Query, Key, and Value matrix as shown in Eq. (\ref{eq:QKV}). The mappings are realized with the feed-forward neural networks. Then, Q and $K^T$ are matrix products and normalized to obtain the attention distribution of each time step. Finally, the attention matrix is superimposed on $V$ to generate an implicit representation with temporal context as Eq. (\ref{eq:attention}).

\begin{equation}
\label{eq:QKV}
Q=\hat{H}W^T_q,K=\hat{H}W^T_k,V=\hat{H}W^T_v
\end{equation}
\begin{equation}
\label{eq:attention}
Attention(Q, K, V)=softmax(\frac{QK^T}{\sqrt{d_k}})V,
\end{equation}
where $W^T_q,W^T_k$, and $W^T_v$ are the projection matrix to be learned, $\sqrt{d_k}$ is the weight scaled factor, and the softmax function is used to normalize the weight scores.

To improve the perception level of temporal semantics in different subspaces, we adopt a multi-head temporal attention module to enrich the representation of information. Multiple sets of self-attentions act independently on the sequence. Their results are concatenated and then linearly transformed to obtain the outputs $H_{Attn}$. The mathematical formula is as follows:
\begin{equation}
\begin{aligned}
&MultiHead(Q, K, V) =Concat(head_1, head_2, \cdots, head_h)W^O,\\
&head_i=Attention(\hat{H}W^Q_i,\hat{H}W^K_i,\hat{H}W^V_i).
\end{aligned}
\end{equation}
where $W^Q_i$, $W^K_i$, and $W^V_i \in 
 \mathbb{R}^{d_{model} \times d_k}$ are all the projection matrices for the linear transformation. $D=h \times d_{model}$, $h$ is the number of heads.

Finally, $H_{Attn}$ is passed into the residual  network, with each sub-layer followed by a Feed-Forward Network and a Layer Normalization. We define the representation of the temporal attention module output as $H_S \in \mathbb{R}^{T \times N \times D}$.

\subsection{Graph Attention Module}
Adaptive graph learner and recurrent graph convolution layer can capture spatial heterogeneity and synchronous spatial-temporal correlations. However, the spatial information of the traffic context is often dynamically evolving \cite{DBLP:conf/cikm/LuGJFZ20}. For example, a major event ({e.g.}, concerts, traffic accidents)  at a location provokes a sudden increase or decrease in vehicles within the neighboring area. To respond to complex and variable spatial patterns, our model applies a graph attention module \cite{velickovic2017graph}. It dynamically assigns different weights based on the feature similarity of the target node and adjacent nodes.

The input to the graph attention module is the set of all node hidden vectors, denoted as $H_{last}=\left\{h_1, h_2, \cdots, h_N \right\}$, where $h_i\in \mathbb{R}^{D}$. Suppose the feature vectors of $v_i$ and $v_j$ be $h_i$ and $h_j$, and $N_i$ be the set of neighbors to $v_i$. The equation to calculate the attention scores between node pairs is as follows:
\begin{equation}
e_{ij}=a(Wh_i,Wh_j),j \in N_i,
\end{equation}
where $(\cdot, \cdot)$ is the concatenation operation, $W \in \mathbb{R}^{F \times D}$ is a learnable linear matrix, $a: \mathbb{R}^{F} \times \mathbb{R}^{F} \rightarrow \mathbb{R}$ as same as $W$ maps the combined parameter matrix into a scalar.

We use the nonlinear activation function LeakyReLU to eliminate minor dependencies. Then the attention scores of $v_i$'s all the adjacent nodes are normalized by the softmax. Formally, the formula is calculated as below:
\begin{equation}
\alpha_{ij}=softmax(e_{ij})=\frac{exp(LeakyReLU(e_{ij}))}{\sum_{k \in N_i}exp(LeakyReLU(e_{ik}))},
\end{equation}

Message aggregation of attention is applied to each node to obtain its output representation. That is:
\begin{equation}
h_i'=\sigma(\sum_{j \in N_i}\alpha_{ij}Wh_j).
\end{equation}
where $\alpha_{ij}$ is the normalized attention score. Generally, we also dropout a certain percentage of it. Here we represent the above steps uniformly in matrix calculations as the following:
\begin{equation}
H_S=ELU \left((M \odot A)H_{last}W\right).
\end{equation}
where $A$ is the adjacency matrix based on the distance relationship. The ELU is also an activation function. The elements in $M\in \mathbb{R}^{N \times N}$ are the dynamic attention factors. $H_{S} \in \mathbb{R} ^{N \times D}$ is the graph embedding of the output in the graph attention module. 

\textbf{The prediction layer} is a typical convolution layer that uses the hidden features of spatial-temporal attention fusion for multi-step prediction. Its input is a graph embedding $H_S$ and temporal tensor $H_T$, which is mapped to the output space  after broadcast summation. Formally, it is as follows:
\begin{equation}
\widehat{X}_{t+1:t+Q} = Conv(H_T + H_S).
\end{equation}
where $\hat{X} \in \mathbb{R}^{Q\times N \times D}$ is the final prediction result of the model.

\section{Experiment}
In this section, we first introduce the experimental settings, including the datasets used, baseline models, and evaluation metrics. We then provide a detailed analysis of the experimental results and visualizations. Finally, we present a series of studies, including module ablation and hyperparameter tuning, to illustrate the effects of model components and parameters on the overall results.

\subsection{Experimental Settings}

{\noindent \bf Dataset.}
We use four traffic flow datasets ({i.e.}, PeMSD3, PeMSD4, PeMSD7, and PeMSD8) collected by the California Department of Transportation sensors on highways \cite{chen2001freeway}. They record traffic statuses such as flow, speed, and occupancy at 5-minute time intervals. The statistical details of the datasets are provided in Table \ref{tab:dataTable}. 

The raw data are standardized using Z-Score \cite{DBLP:conf/aaai/SongLGW20} and then divided into the training set, validation set, and test set with a ratio of 6: 2: 2. The training set is disrupted before training and the validation set is used to control the early stopping of the training process. In the multi-step prediction, we set both the input sequence $P$ and the output sequence $Q$ to 12.

\begin{table}
  \caption{The statistics of the tested real-world datasets.}
  \label{tab:dataTable}
  \centering
  \resizebox{1.01\linewidth}{!}{
  \begin{tabular}{ccccc}
    \toprule
    Dataset & Sensors & Edges & Samples & Time Range\\
    \midrule
    PeMSD3 & 358 & 547 & 26,208 & Sept - Nov, 2018 \\
    PeMSD4 & 307 & 340 & 16,992 & Jan - Feb, 2018  \\
    PeMSD7 & 883 & 866 & 28,224 & May - Aug, 2017  \\
    PeMSD8 & 170 & 296 & 17,856 & Jul - Aug, 2016  \\
  \bottomrule
\end{tabular}
}
\end{table}

{\noindent \bf Baseline Methods.} To evaluate the proposed model, we compare it with 18 baseline models, including traditional statistical, deep learning, and graph neural network models. These baselines serve as a benchmark to demonstrate the effectiveness of our approach. Below is a brief overview of the critical baseline:

\begin{itemize}
\vspace{-2.0mm}
\item \textbf{VAR}  \cite{DBLP:journals/tits/WuHL04}: Vector Auto-Regression assumes that the historical time series is stationary and forecasts by estimating the relationship between the time series and its lagged values.
\item \textbf{ARIMA} \cite{kumar2015short}: Auto-regressive Integrated Moving Average with Kalman filter is a widely used statistical model for time series.
\item \textbf{FC-LSTM} \cite{sutskever2014sequence}: A recurrent neural network with fully connected LSTM hidden units.
\item \textbf{STGCN} \cite{DBLP:conf/ijcai/YuYZ18}: Spatial-Temporal Graph Convolutional Network, which com-
bines graph convolution with 1D gated convolution.
\item \textbf{DCRNN} \cite{DBLP:conf/iclr/LiYS018}: Diffusion convolution recurrent neural network, which combines graph convolution networks with recurrent neural networks in an encoder-
decoder architecture.
\item \textbf{GWNET} \cite{wu2019graph}: Graph WaveNet introduces an adaptive adjacency matrix and combines diffuse graph convolution with TCN instead of 1D convolution.
\item \textbf{AGCRN} \cite{DBLP:conf/nips/0001YL0020}: Adaptive Graph Convolutional Recurrent Network, which augments traditional graph convolution with adaptive graph generation and node adaptive parameter learning, and is integrated into a recurrent neural network to capture more complex spatial-temporal correlations.
\item \textbf{STG-NCDE} \cite{choi2022graph}: Spatial-Temporal Graph Neural Controlled Differential Equation, which designs two NCDEs for temporal processing and spatial processing and integrates them into a single framework.
\item \textbf{DSTAGNN} \cite{lan2022dstagnn}: Dynamic Spatial-Temporal Aware Graph Neural Network, which captures dynamic spatial and temporal dependencies between nodes and receptive field features through improved multi-headed attention and multi-scale gated convolution.

\vspace{-2.0mm}
\end{itemize}
{\noindent \bf Evaluation Metrics.} Throughout the experiments, we employ Mean Absolute Error (MAE), Root Mean Squared Error (RMSE), and Mean Absolute Percentage Error (MAPE)  \cite{DBLP:conf/www/Wang0WJWTJY20} as evaluation metrics to compare the overall performance of the different models on the test dataset.  A lower MAE, RMSE, and MAPE value indicates a better prediction performance.

{\noindent \bf Parameters Setup.} Our task is to predict the traffic state in the next hour based on the historical traffic data of the previous hour. In our experiments, we set all the hidden dimensions to 64, the head of attention to 4, and the order of graph convolution to 2. We use the Adam \cite{DBLP:journals/corr/KingmaB14} optimizer for training the model. We select SmoothL1Loss as the loss function. The batch size of the data is 64, the number of epochs is 300, and the initialized learning rate is 0.003. An early stopping strategy is used with a patience of 15 iterations on the validation dataset to prevent overfitting.

\subsection{Experimental Results}
Table \ref{tab:results} presents the evaluation results of AFDGCN and major baseline methods on the four tested datasets. The bold values and the underlined values are the best and second-best prediction performances, respectively. Our method (AFDGCN) always achieves the best performance and is in bold. Compared with the best baseline (underlined), the performance of AFDGCN in four datasets (PeMSD3, PeMSD4, PeMSD7, and PeMSD8) is improved by 3.85\%, 0.62\%, 1.51\%, and 2.78\% on MAE and 3.41\%, 0.63\%, 3.50\%, and 2.42\% on MAPE respectively. In particular, the improvements are more significant for the PeMSD3 and PeMSD8 datasets. 

\begin{table*}[t]
 \renewcommand{\arraystretch}{1.0}
 \centering
 \caption{Performance comparison of AFDGCN and other baseline models. AFDGCN achieves the best performance for all datasets.}
 \vspace{-2mm}
 \resizebox{1.01 \linewidth}{!}{
 \begin{tabular}{c|c c c|c c c |c c c|c c c}
  \toprule[1.2pt]
  \multirow{2}{*}{Model} & \multicolumn{3}{c|}{PEMSD3} & \multicolumn{3}{c|}{PEMSD4} & \multicolumn{3}{c|}{PEMSD7} & \multicolumn{3}{c}{PEMSD8} \\
  \cline{2-13}
  {}& MAE & RMSE & MAPE & MAE & RMSE & MAPE & MAE & RMSE & MAPE & MAE & RMSE & MAPE\\
  \midrule
  HA & 31.58 & 52.39 & 33.78\% & 38.03 & 59.24 & 27.88\% & 45.12 & 65.64 & 24.51\% & 34.86 & 59.24 & 27.88\% \\
  ARIMA & 35.41 & 47.59 & 33.78\% & 33.73 & 48.80 & 24.18\% & 38.17 & 59.27 & 19.46\% & 31.09 & 44.32 & 22.73\% \\
  VAR & 23.65 & 38.26 & 24.51\% & 24.54 & 38.61 & 17.24\% & 50.22 & 75.63 & 32.22\% & 19.19 & 29.81 & 13.10\%  \\ 
  FC-LSTM & 21.33 & 35.11 & 23.33\% & 26.77 & 40.65 & 18.23\% & 29.98 & 45.94 & 13.20\% & 23.09 & 35.17 & 14.99\% \\ 
  TCN & 19.32 & 33.55 & 19.93\% & 23.22    & 37.26  & 15.59\%  & 32.72 & 42.23 & 14.26\%  & 22.72  & 35.79 & 14.03\%  \\
  STGCN & 17.55 & 30.42  & 17.34\% & 21.16  & 34.89  & 13.83\%  & 25.33 & 39.34 & 11.21\%  & 17.50  & 27.09  & 11.29\% \\ 
  DCRNN  & 17.99  & 30.31  & 18.34\% & 21.22  & 33.44  & 14.17\% & 25.22  & 38.61  & 11.82\% & 16.82  & 26.36  & 10.92\%  \\ 
  GWNET & 19.12  & 32.77  & 18.89\% & 24.89  & 39.66  & 17.29\%  & 26.39 & 41.50  & 11.97\%  & 18.28    & 30.05 & 12.15\% \\ 
  STG2Seq  & 19.03 & 29.83 & 21.55\% & 25.20  & 38.86 & 13.18\%  & 32.77    & 47.16 & 20.16\%  & 20.17    & 30.71 & 17.32\%  \\
  LSGCN   & 17.94  & 29.85  & 16.98\% & 21.53  & 33.86  & 13.18\%  & 27.31   & 41.46  & 11.98\%  & 17.73  & 26.76   & 11.20\%  \\
  ASTGCN  & 17.34  & 29.56  & 17.21\%  & 22.93   & 35.22  & 16.56\%  & 24.01  & 37.87   & 10.73\%  & 18.25  & 28.06  & 11.64\%  \\ 
  STSGCN  & 17.48  & 29.21  & 16.78\%  & 21.19  & 33.65 & 13.90\%  & 24.26  & 39.03  & 10.21\%  & 17.13 & 26.80  & 10.96\% \\ 
  STFGNN  & 16.77  & 28.34  & 16.30\% & 20.48   & 32.51  & 16.77\%  & 23.46  & 36.60  & 9.21\%  & 16.94  & 26.25  & 10.60\%  \\
  STGODE  & 16.50 & 27.84  & 16.69\% & 20.84  & 32.82   & 13.77\% & 22.59  & 37.54  & 10.14\% & 16.81  & 25.97   & 10.62\% \\ 
  AGCRN  & 15.98  & 28.25  & 15.23\% & 19.83  & 32.26  & 12.97\%  & 22.37 & 36.55  & 9.12\% & 15.95  & 25.22  & 10.09\% \\
  Z-GCNETs  & 16.64  & 28.15 & 16.39\% & 19.50  & 31.61 & 12.78\%  & 21.77  & 35.17  & 9.25\%  & 15.67  & 25.11  & 10.01\%  \\ 
  STG-NCDE  & \underline{15.57} & \underline{27.09} & 15.06\% & \underline{19.21} & \underline{31.09} & 12.76\% & \underline{20.53} & \underline{33.84} & \underline{8.80\%} & \underline{15.45} & 24.81  & \underline{9.92\%}  \\ 
  DSTAGNN &  \underline{15.57} & 27.21 & \underline{14.68\%} & 19.30  & 31.46  & \underline{12.70\%} & 21.67 & 34.51  & 9.01\%  & 15.67   & \underline{24.77} & 9.94\%  \\
\midrule
  \textbf{AFDGCN}        & \textbf{14.97} & \textbf{25.81} & \textbf{14.18\%} & \textbf{19.09} & \textbf{31.01} & \textbf{12.62\%} & \textbf{20.22} & \textbf{33.80} & \textbf{8.52\%} & \textbf{15.02} & \textbf{24.37} & \textbf{9.68\%} \\   \bottomrule[1.2pt]
 \end{tabular}
 }
 \label{tab:results}
\end{table*}

\begin{figure}[!t]
\renewcommand{\arraystretch}{1.0}
\centering
\begin{subfigure}{\linewidth}
\centering
\includegraphics[width=0.50\textwidth]{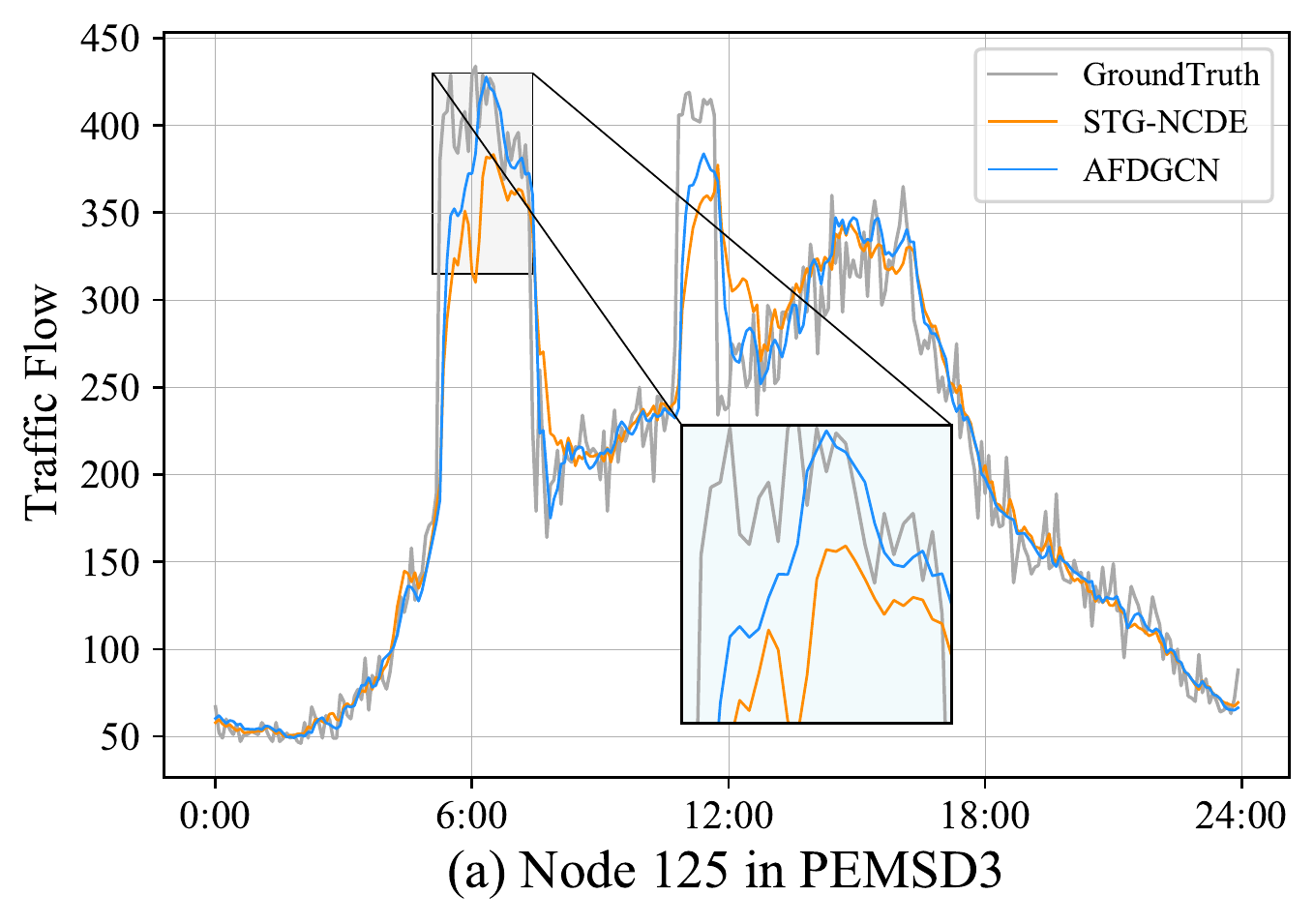}
\hspace{-2mm}
\includegraphics[width=0.50\textwidth]{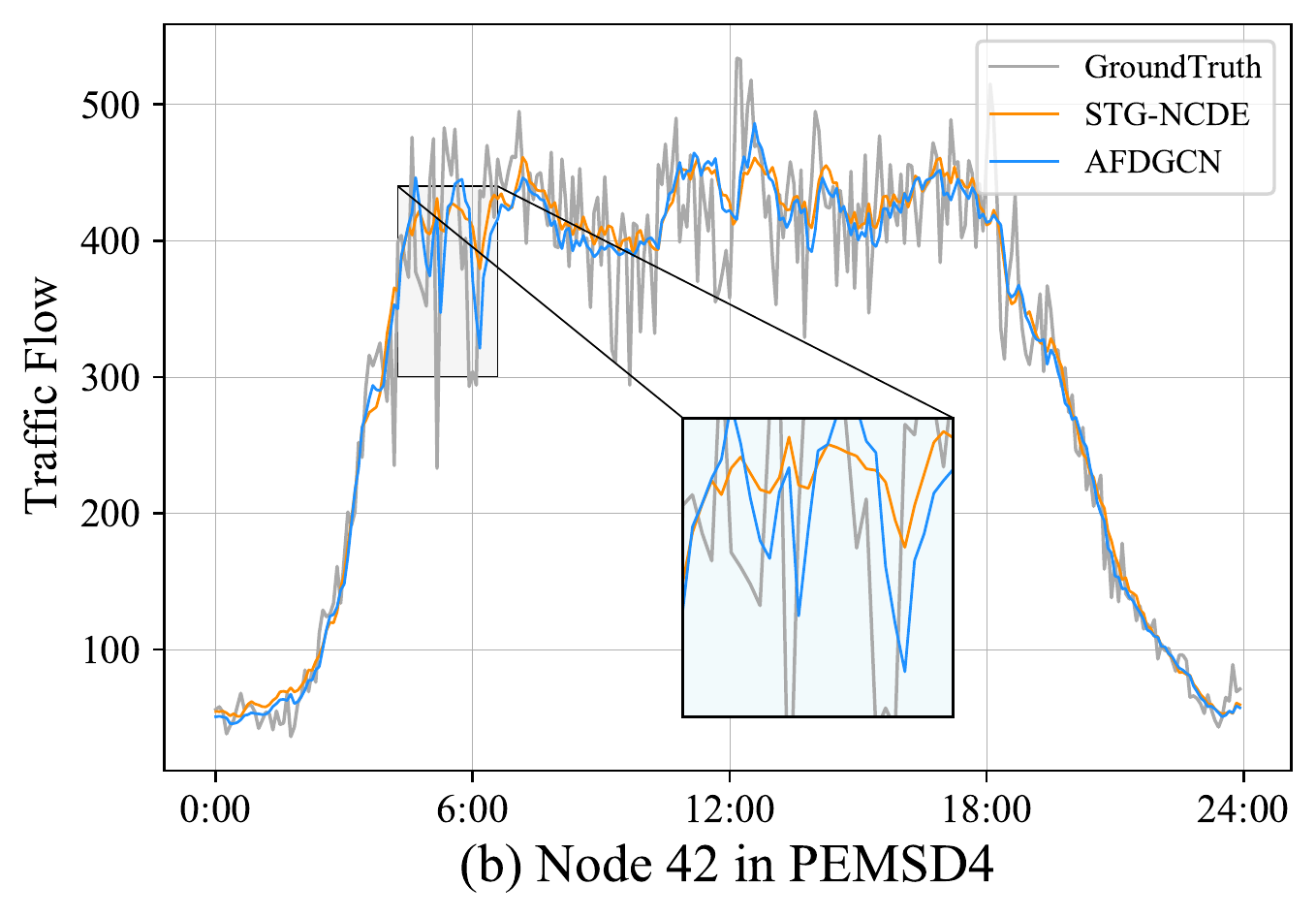}
\end{subfigure}

\begin{subfigure}{\linewidth}
\centering
\includegraphics[width=0.50\textwidth]{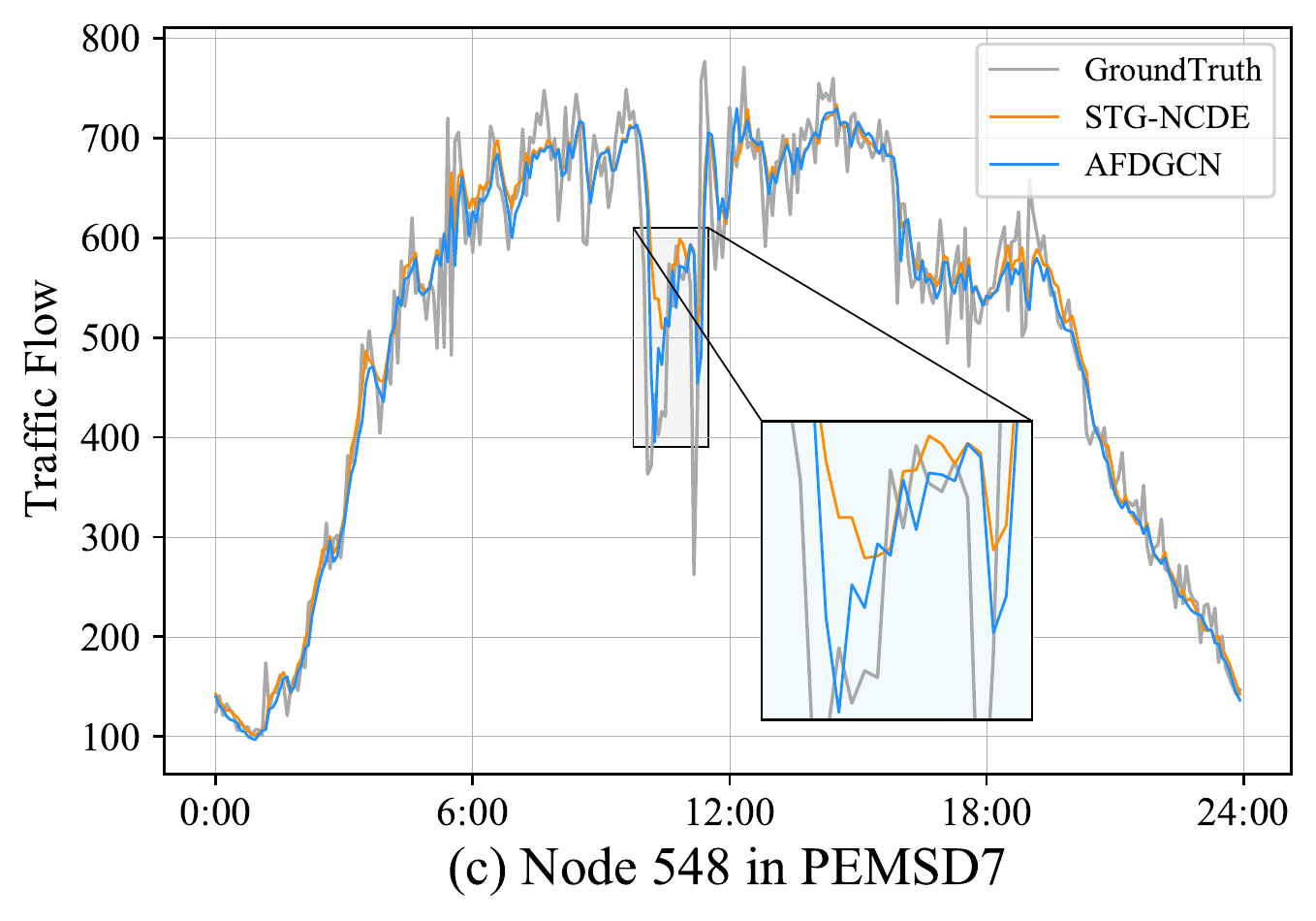}
\hspace{-2mm}
\includegraphics[width=0.50\textwidth]{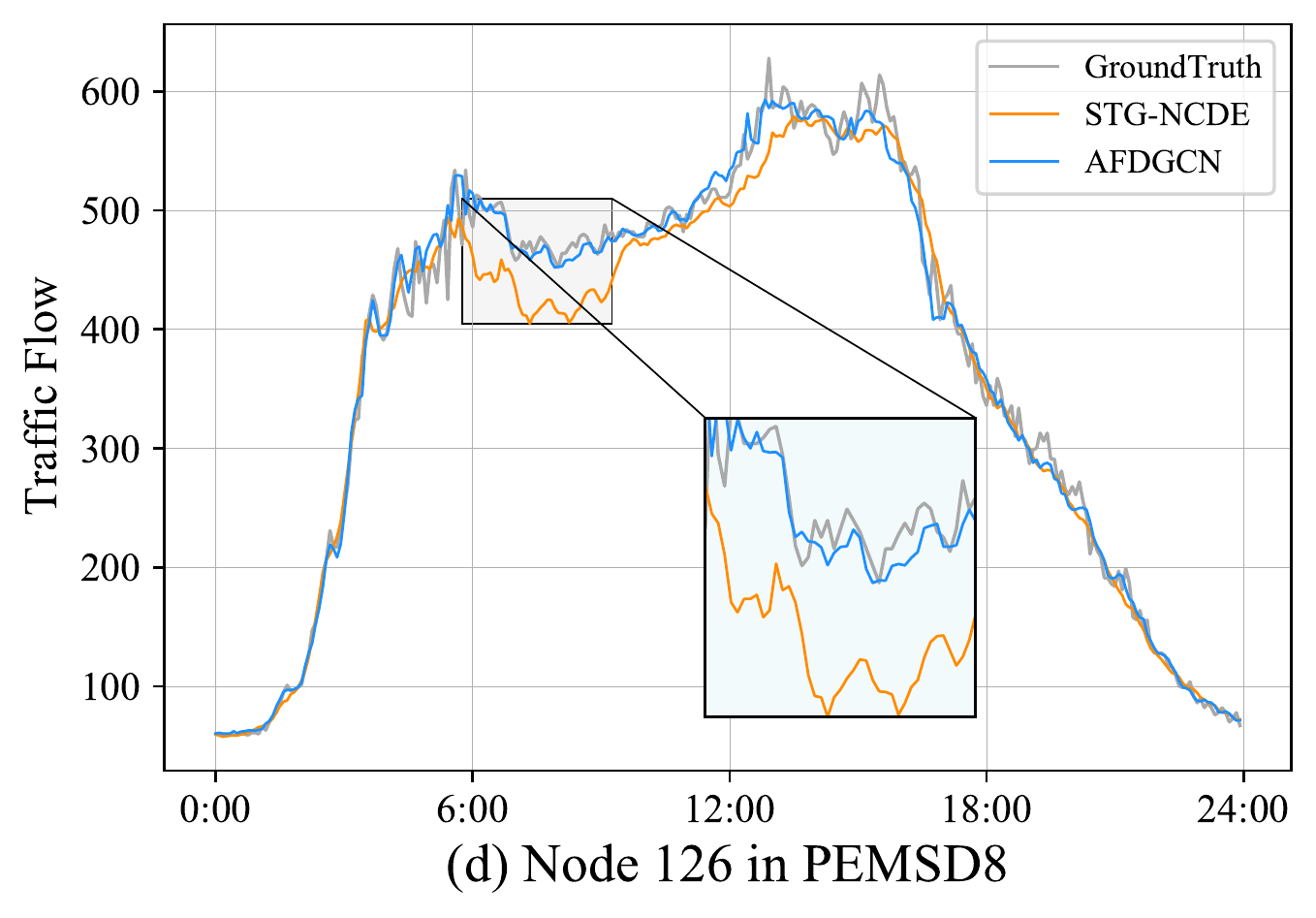}
\end{subfigure}
\vspace{-6mm}
\caption{Traffic forecasting within one day in different datasets.}
\label{fig:subfigure-grid}
\end{figure}

We observe that statistical methods such as HA, ARIMA, and VAR have difficulty in handling nonlinear, non-stationary time series data and the model prediction errors are high. Although deep learning methods such as FC-LSTM and TCN have advantages over statistical models, they only consider temporal correlations and cannot exploit spatial dependencies, thus having a limited ability to model spatial-temporal data. Spatial-temporal graph networks such as STGCN \cite{yao2019revisiting}, DCRNN \cite{DBLP:conf/iclr/LiYS018}, and GWNET \cite{wu2019graph} are designed with spatial-temporal components, so they generally have better performance compared with temporal-only-based methods. AGCRN and DSTAGNN are free from the limitation of fixed graph structure and use self-adaptive or dynamic properties of spatial association between nodes in historical data to construct the graph structure. They further improve the prediction performance compared with the previous methods. In contrast, our method not only considers the effects of synchronous spatial-temporal correlation but also mines the multi-scale and long-distance traffic fluctuation relationships from the perspective of spatial-temporal attention.

Furthermore, we quantify the difference between our model and the best baseline model using the relative error rate. Across the four datasets, the average MAE, RMSE, and MAPE values for AFDGCN are 17.33, 28.75, and 11.25\%, respectively; the corresponding average values of DSTAGNN \cite{lan2022dstagnn} are 18.05 (104.2\%), 29.48 (102.5\%), and 11.58\% (103.0\%), and the average error performance of STG-NCDE \cite{choi2022graph} are 17.69 (102.1\%), 29.21 (101.6\%) and 11.64\% (103.5\%). Taken together, the improved gain of our model compared to these two models ranges from 1.6\% to 4.2\%.

To clearly examine the difference in prediction performance between our method and the baseline methods, we plot in Fig. \ref{fig:subfigure-grid} the predicted and ground truth of our method (AFDGCN) versus STG-NCDE for several stations/nodes within a given day. AFDGCN and STG-NCDE well fit the real conditions of traffic variations in many time periods, but our model shows more accurate prediction performance in challenging traffic scenarios, {e.g.}, during traffic peak hours or when the magnitude of fluctuations is drastic. In particular, our method adapts well to changes in traffic trends, {e.g.}, 5:00-8:00 in (a), 4:00-7:00 in (b), 10:00-12:00 in (c), and 6:00-9:00 in (d). In contrast, the prediction curves of STG-NCDE significantly deviate from the ground truth because of its limited predictive power. 

\subsection{Additional Experimental Studies}
{\noindent \bf Efficiency study.} We first compare the computational cost of our model AFDGCN with several competitive baseline models in the PEMSD4 using the same Graphic Card NVIDIA Tesla V100. As shown in Table \ref{tab:Efficiency}, AFDGCN has a comparable training and inference time as AGCRN but attains much better predictive performance. Compared with STGODE, the training time and inference time of AFDGCN are reduced by 56.13\% and 64.10\%. Although STG-NCDE has fewer parameters than AFDGCN, the complex model structure and operators make the computational cost of STG-NCDE significantly higher than AFDGCN. The training time and inference time of AFDGCN are 4-5x and 4-6x faster than DSTAGNN and STG-NCDE, respectively. Remarkably, our model outperforms all the baseline methods in the predictive performance by only using highly competitive training and inference time.

\begin{figure}[!t]
\renewcommand{\arraystretch}{1.0}
\centering
\begin{subfigure}{\linewidth}
\centering
\includegraphics[width=0.50\textwidth]{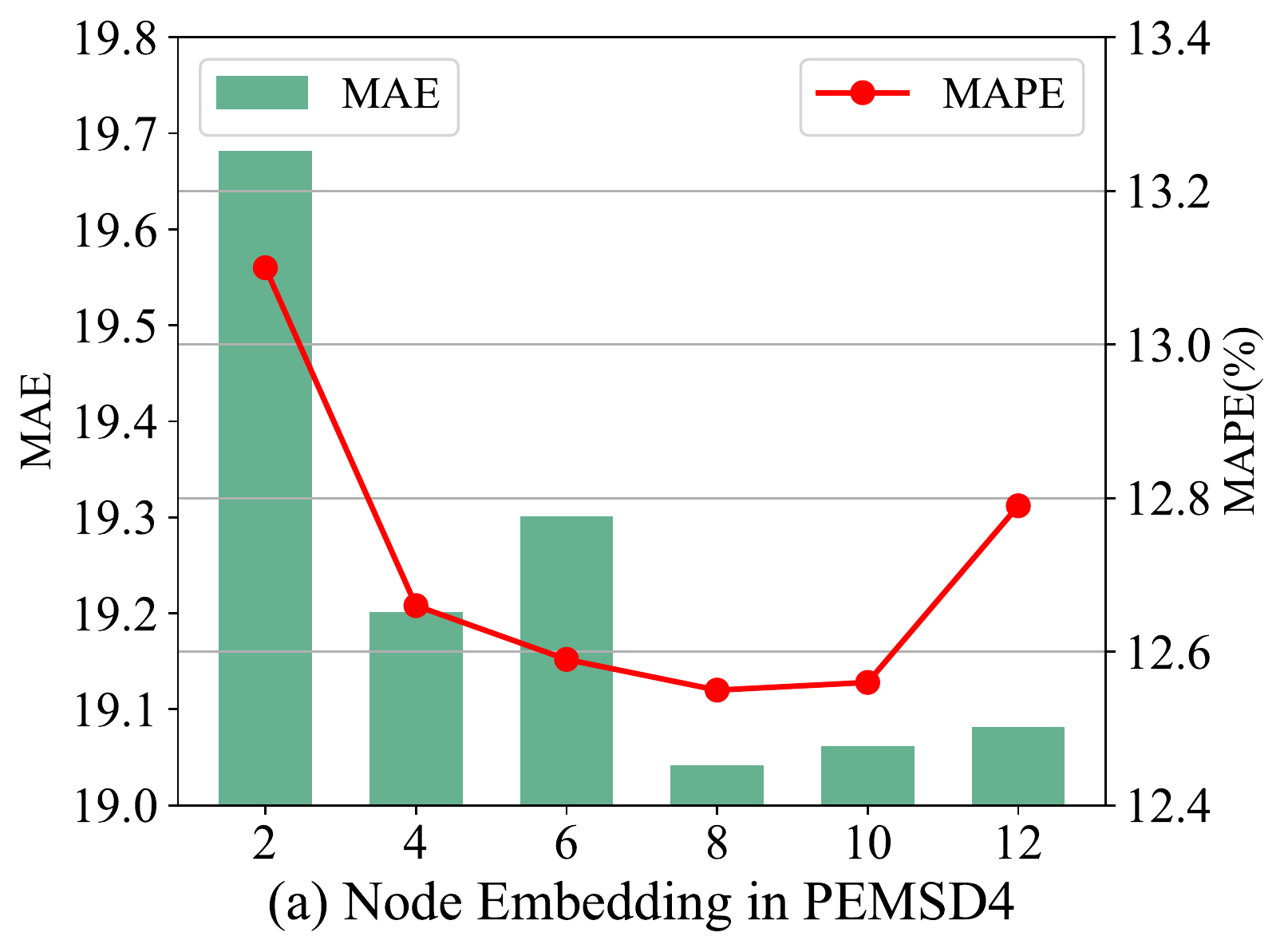}
\hspace{-2mm}
\includegraphics[width=0.50\textwidth]{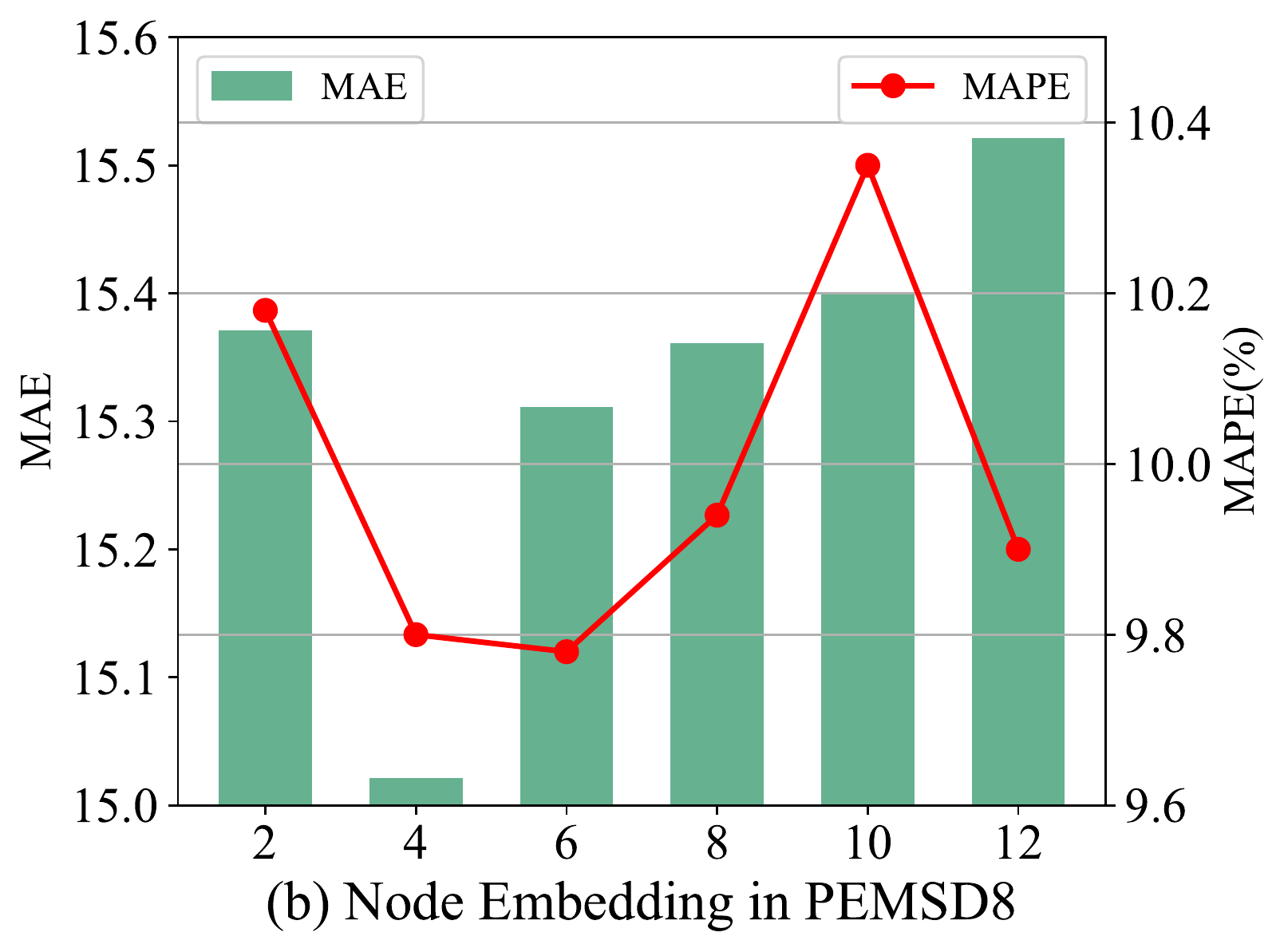}
\end{subfigure}
\caption{The effects of the Node Embedding Dimension.}
\label{fig:subfigures}
\end{figure}

{\noindent \bf Ablation study.} We perform an ablation study on different modules of our model to investigate their individual contributions in the PeMSD4 and PeMSD8. Let \textbf{DGCGRU} denotes the dynamic graph convolution recurrent network, and \textbf{DGCGRU with Attn} denote DGCGRU with multi-head temporal attention. We denote the full model AFDGCN with the feature augmentation removed and AFDGCN with the graph attention layer removed as \textbf{w/o FAL} and \textbf{w/o GAT}, respectively. The prediction results of these methods for PeMSD4 and PeMSD8 are summarized in Table \ref{tab:Ablation}. DGCGRU performs normally as expected. After adding the multi-headed temporal attention mechanism, DGCGRU with Attn achieves significant improvements in the three evaluation metrics. This demonstrates the effectiveness of the temporal attention mechanism in modeling global temporal dependence. In addition, AFDGCN with either the feature augmentation layer or the graph attention layer removed has a noticeable loss in the prediction performance.

{\noindent \bf Hyperparametric study.} An essential parameter in AFDGCN is the node embedding size, which not only affects the quality of the learning graph but also determines the diversity of parameters in the DGCGRU layer. Fig. \ref{fig:subfigures} show the effects of different node embedding dimensions on the model prediction results. It can be found that the optimal node embedding dimensions for the PeMSD4 and PeMSD8 datasets are 8 and 4, respectively. On the one hand, node embedding with a larger value contains more parameter information, which improves the expressiveness of the model to infer more complete spatial correlations. On the other hand, the larger the number of parameters, the more likely the model is prone to overfitting.

\begin{figure}[!t]
\centering
\includegraphics[width=\linewidth]{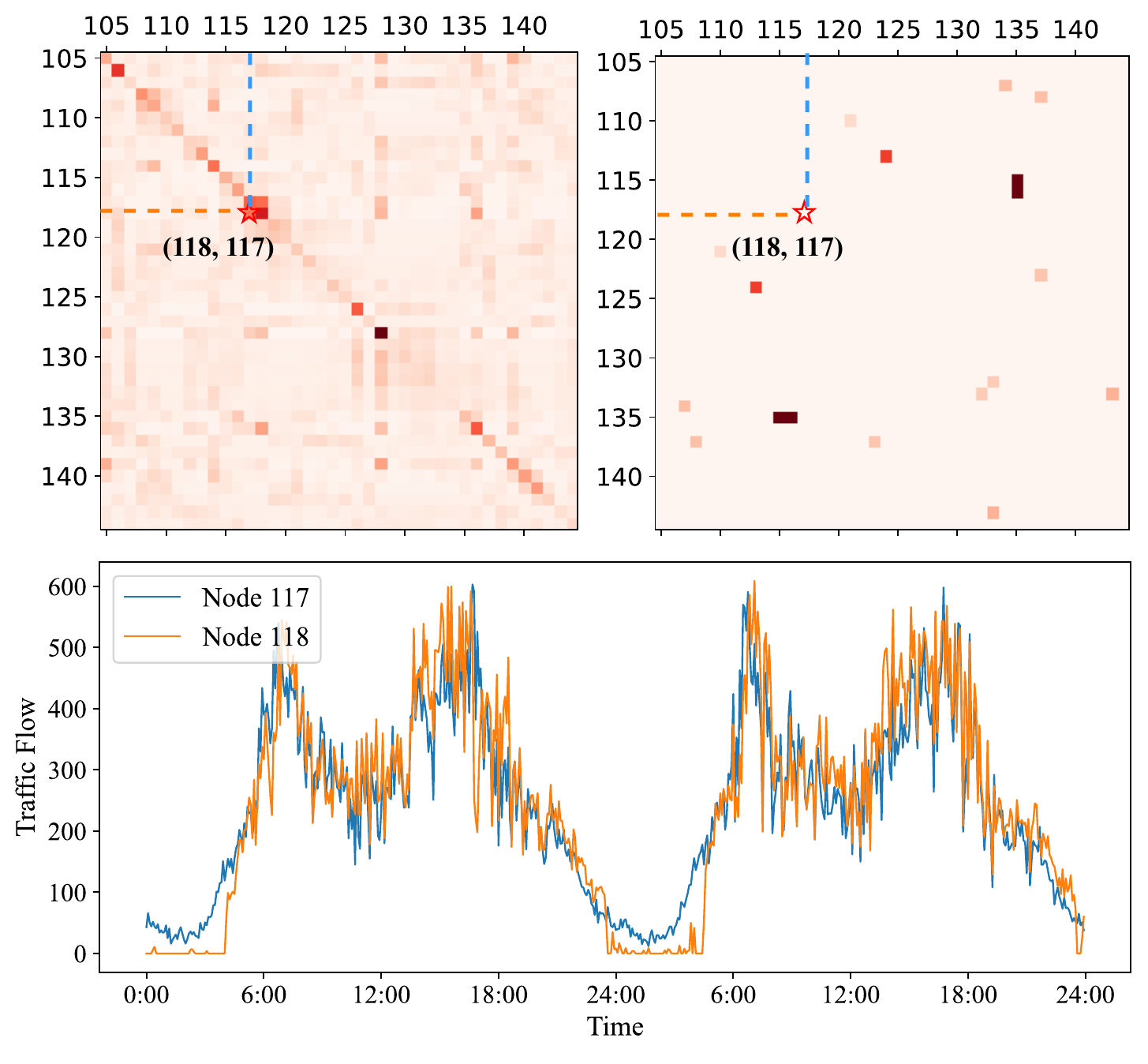}
\vspace{-0.15in}
\caption{The heatmap visualization for dynamic generation graph (Left) and pre-defined graph (Right) in PeMSD4.}
\label{fig:label10}
\vspace{-0.10in}
\end{figure}

{\noindent \bf Interpretability study.} Now we provide a shred of evidence on the benefit of using an adaptive graph structure. We visualize the correlation between some node pairs under adaptive and pre-defined graphs using heatmaps in PeMSD4, as shown in Fig. \ref{fig:label10}. The darker the color, the stronger the correlation between the nodes. The pre-defined graph (right) relies only on geographic distance, and the information presented by the heatmap is sparse and single. In contrast, the information in the dynamically generated graph (left) learned with parameters is dense and diverse. We take two points in the road network as an example (as shown by the star in the figure), which do not exhibit correlation in the pre-defined static graph. But in fact, the two points (Node 117, Node 118) are very close to each other in terms of the period and trend changes of traffic as in the bottom figure. Therefore, the dynamic adjacency matrix can implicitly learn the dynamic representation of the road network and provide an effective complement to the static adjacency matrix.

\begin{table}
\renewcommand{\arraystretch}{1.02}
\centering
\caption{The training and inference time in PEMSD4 dataset under a Tesla V100 GPU(s/epoch).}
\vspace{-2mm}
\label{tab:Efficiency}
{\fontsize{3.2}{4.}\selectfont
\resizebox{.99\linewidth}{!}{
\begin{tabular}{cccc}
\specialrule{0.15em}{0.1em}{0.4em}
Model   & \#Params & \#Training & \#Inference \\ \specialrule{0.05em}{0.1em}{0.4em}
DSTAGNN  & 3,579,728    & 134.72            & 14.94        \\ 
STG-NCDE & 322,588     & 92.04             & 10.94        \\ 
STGODE   & 714,504     & 57.38             & 7.27         \\ 
AGCRN    & 748,810     & 21.68             & 2.88         \\ 
AFDGCN   & 435,121     & 25.17             & 2.61         \\ \specialrule{0.15em}{0.1em}{0.4em}
\end{tabular}
}
}
\vspace{-2mm}
\end{table}

\begin{table}
 \renewcommand{\arraystretch}{1.2}
 \centering
 \vspace{-3mm}
  \caption{The results of an ablation study in AFDGCN.}
   \vspace{-2mm}
 \resizebox{1.00\linewidth}{!}{
 \begin{tabular}{c|c c c|c c c }
  \toprule[1.2pt]
  \multirow{2}{*}{Module} & \multicolumn{3}{c|}{PEMSD4} &  \multicolumn{3}{c}{PEMSD8} \\
  \cline{2-7}
  {}& MAE & RMSE & MAPE & MAE & RMSE & MAPE \\
  \midrule
  DGCGRU & 19.80 & 32.20 & 13.18\% &  16.01 &  25.51 & 10.35\% \\
  DGCGRU with Attn & 19.32 & 31.87 &  12.79\% &   15.53 &   24.91 &  10.01\% \\
  AFDGCN w/o FAL & 19.11 & 31.65 &  12.67\% &   15.35 &   24.49 &  9.98\% \\
  AFDGCN w/o GAT & 19.11 & 31.48 &  12.61\% &   15.29 &   24.68&  9.71\% \\
  AFDGCN & 19.04 & 31.11 &  12.55\% &   15.02 &   24.36 &  9.80\% \\
  \bottomrule[1.2pt]
 \end{tabular}
 }
 \label{tab:Ablation}
 \vspace{-3mm}
\end{table}

\section{Conclusion}
In this paper, we propose a novel spatial-temporal graph neural network for traffic prediction. The model enhances the traditional GCN by adopting a dynamic generation graph with node parametric learning, and combines the improved GCN with GRU to capture synchronous spatial-temporal correlation. To handle long-range and multi-view changes in complex traffic scenes, we introduce a spatial-temporal attention fusion module to effectively improve the model's performance. We conduct experiments on four publicly available traffic datasets, and the results demonstrate the effectiveness and superiority of AFDGCN. In future studies, we plan to extend AFDGCN to other spatial-temporal prediction tasks and explore modeling dynamic spatial dependencies with time-varying properties.

\ack Detian Zhang is supported by the Collaborative Innovation Center of Novel Software Technology and Industrialization, the Priority Academic Program Development of Jiangsu Higher Education Institutions. Chunjiang Zhu is supported by UNCG Start-up Funds and Faculty First Award. 

\bibliography{ecai}
\end{document}